\documentclass[sigconf]{acmart}

\usepackage{algorithmic}
\usepackage{graphicx}
\usepackage{comment}
\usepackage{textcomp}
\usepackage{xcolor}
\usepackage{subcaption}

\def\BibTeX{{\rm B\kern-.05em{\sc i\kern-.025em b}\kern-.08em
    T\kern-.1667em\lower.7ex\hbox{E}\kern-.125emX}}

\setcopyright{none}

\begin{document}
\settopmatter{printacmref=false}
{\fontfamily{phv}\selectfont
\title{3D object reconstruction and 6D-pose estimation from 2D shape for robotic grasping of objects}
\settopmatter{printacmref=false}

\author{Marcell Wolnitza}
\authornote{The author is also with Georg-August-University G\"ottingen, 37077 G\"ottingen, Germany.}
\email{wolnitza@hs-koblenz.de}
\affiliation{%
  \centering
  \institution{\mbox{University~of~Applied~Sciences~Koblenz}}
  \institution{\mbox{Faculty~of~Mathematics~and~Technology}}
  \institution{53424 Remagen, Germany}
  \Large
  }

\author{Osman Kaya}
\email{osman.kaya@uni-goettingen.de}
\affiliation{%
  \centering
  \institution{Georg-August-University~G\"ottingen}
  \institution{Third Institute of Physics - Biophysics}
  \institution{37077 G\"ottingen, Germany}
  \Large
  }
  
\author{Tomas Kulvicius}
\authornote{The author is also with the University Medical Center Göttingen,\linebreak
Child and Adolescent Psychiatry and Psychotherapy, 37077 G\"ottingen, Germany.}
\email{tomas.kulvicius@uni-goettingen.de}
\affiliation{%
  \centering
  \institution{Georg-August-University~G\"ottingen}
  \institution{Third Institute of Physics - Biophysics}
  \institution{37077 G\"ottingen, Germany}
  \Large
  }
  
\author{Florentin Wörgötter}
\authornote{We acknowledge funding by DFG WO388 /1-16 to F.W.}
\email{worgott@gwdg.de}
\affiliation{%
  \centering
  \institution{Georg-August-University~G\"ottingen}
  \institution{Third Institute of Physics - Biophysics}
  \institution{37077 G\"ottingen, Germany}
  \Large
  }

\author{Babette Dellen}
\email{dellen@hs-koblenz.de}
\affiliation{%
  \centering
  \institution{\mbox{University~of~Applied~Sciences~Koblenz}}
  \institution{\mbox{Faculty~of~Mathematics~and~Technology}}
  \institution{53424 Remagen, Germany}
  \Large
  }


%
%
%
%
%
\renewcommand{\shortauthors}{ }
\renewcommand{\footnotesize}{\normalsize}
\renewcommand{\arraystretch}{1.3}
}
\begin{abstract}
We propose a method for 3D object reconstruction and 6D-pose estimation from 2D images that uses knowledge about object shape as the primary key. In the proposed pipeline, recognition and labeling of objects in 2D images deliver 2D segment silhouettes that are compared with the 2D silhouettes of projections obtained from various views of a 3D model representing the recognized object class. By computing transformation parameters directly from the 2D images, the number of free parameters required during the registration process is reduced, making the approach feasible. Furthermore, 3D transformations and projective geometry are employed to arrive at a full 3D reconstruction of the object in camera space using a calibrated set up. Inclusion of a second camera allows resolving remaining ambiguities. The method is quantitatively evaluated using synthetic data and tested with real data, and additional results for the well-known Linemod data set are shown. In robot experiments, successful grasping of objects demonstrates its usability in real-world environments, and, where possible, a comparison with other methods is provided. The method is applicable to scenarios where 3D object models, e.g., CAD-models or point clouds, are available and precise pixel-wise segmentation maps of 2D images can be obtained. Different from other methods, the method does not use 3D depth for training, widening the domain of application.  
\end{abstract}
\maketitle

\section{Introduction}
\label{intro}
Object knowledge plays an important role in 3D visual perception. For example, an object located at a large distance from an observer will cover only a very small retinal area or, in case of a camera, the 2D image. The size of this area provides valuable information about the distance to the object \cite{sousa2011}. The shape of the silhouette and its position and orientation in the 2D image provide further information about object pose and 3D structure. These cues are potentially equally important as stereo, motion, shading and texture cues \cite{sousa2011,howard2012,eimer1996}. 
\begin{figure}
    \centering
    \includegraphics[width=0.85\linewidth]{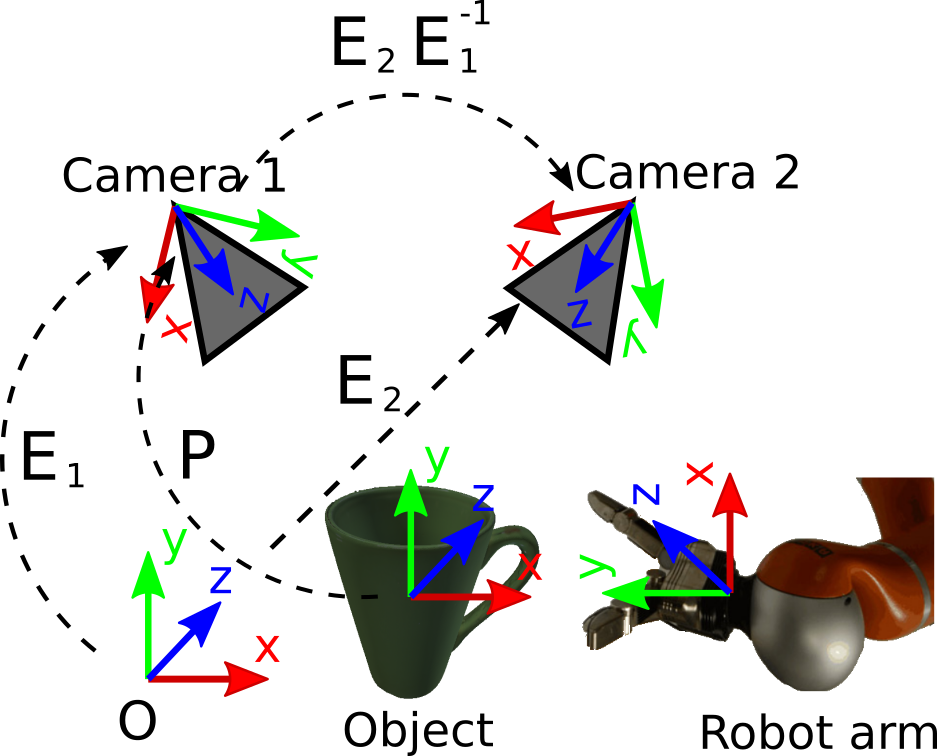}
    \caption{Overview of the experimental setup and the coordinate systems involved. The goal of our method is to find the transformation matrix $P$, which describes the object in the reference system of camera 1. This leads to a full 3D reconstruction of the object.}
    \label{fig:schematic}
\end{figure}
However, exploiting the 2D object shape for extracting 3D information requires both a precise segmentation of the 2D image and accurate recognition of the object class. Only until recently, due to advances of deep-learning approaches in the field object classification and segmentation \cite{unet, fcnn, maskrcnn}, this problem can be tackled in a controlled environment, if sufficient training data is available. Based on this idea, we developed a computer-vision pipeline for estimating the 6D object pose and reconstructing the 3D point cloud of the object from a single or a pair of views using object knowledge explicitly as the main cue. The pipeline follows mostly a classical, hierarchical computer-vision approach. A deep-learning method based on ResNet50V2 \cite{resnet} is only employed at the beginning to recognize and segment the objects from the 2D image. Then, the 2D shape of the image segment is compared with model shapes taken from an object database and the six pose parameters are estimated. Together with the intrinsic and extrinsic matrices of the calibrated set up, a full reconstruction of the object is achieved by transforming the model point cloud into the camera space. 

For the robotic experiments, a set of real objects was selected and their respective 3D models (point clouds) were acquired with a 3D scanner. Real images of the objects placed in a scene were taken with a calibrated set up composed of two cameras and a robotic arm (see Fig.~\ref{fig:schematic}). Synthetic test images were generated with the software Blender \cite{blender} using the 3D models of the objects and simulating the cameras of the real set up. This way, the segmentation method could be trained on a large data set resembling the real-world scenario. Using manually-segmented real images from the scene, further training and fine tuning of the segmentation method could be achieved. The segmentation maps obtained for new images of the scene were precise enough to extract pose from segment shape and conducting robotic grasping experiments. A successful grasp indicated a correct 3D reconstruction within the error tolerance of the robotic grasp motion. The success rates could then be compared to results for the method DOPE reported in \cite{dope}.

The synthetic data described above was also used to benchmark the method. To provide a further comparison with other methods, the method was also applied to the Linemod data set of the BOP challenge \cite{hodan2018bop,hodan2020bop}. However, the data sets of the BOP challenge are not suitable for testing our method, because most of the high-ranking deep-learning-based pose-estimation methods, e.g., \cite{posecnn,dope}, have other requirements on the data than our method. This will be detailed in Section~\ref{sec:linemod} and \ref{discussion}.

The main contributions of this work can be summarized as follows: (i) A framework for 6D-pose estimation and 3D object reconstruction from single or pairs of RGB images is provided. The feasibility of the approach is demonstrated using robotic grasping experiments in a real-world scenario. (ii) Except for the image-segmentation front end, the method follows a purely classical approach, and as such does not require training with 3D data or 6D poses, which are usually difficult to acquire in real-world set ups. (iii) The shape of an  object can be considered as an universal cue to object pose and is independent of individual appearances in terms of color or texture, which could prove beneficial in the future. (iv) A data set including two camera views of objects together with their 3D object models and intrinsic and extrinsic camera parameters has been acquired and will be made available with the paper.      

\section{Related work}
\label{related_work}
\subsection{2D shapes for 3D reconstruction}
2D object shapes have been previously used to extract 3D information from images \cite{walck2010automatic,dellen2009}, e.g., to compute disparities by comparing silhouettes in stereo images \cite{dellen2009}, or to apply methods related to volume-carving techniques \cite{walck2010automatic,bone2008automated}. In \cite{walck2010automatic}, an approach for automatic object reconstruction of unknown objects was developed that uses object silhouettes from different viewpoints for reconstruction and estimation of object distance. However, different from our approach, no object models were used. Furthermore, segmentation was performed using a semi-manual approach based on graph cuts including online feedback from the robot to calculate visual features, while in our approach segmentation is automatic and no robot feedback is required. In our method, semi-manual labeling is only needed for generating training data for the image-segmentation front end, but not in the final application. In \cite{bone2008automated}, a camera mounted on a robot arm was used to record object shapes online to generate a 3D surface model of unknown objects. The method relied on an additional laser scanner mounted on the robot arm to refine the surface model, which makes the method very hardware dependent. A grasp planning algorithm was finally used for picking objects. Different from our approach, no object models were employed here. 
Using a priori object models for 3D reconstruction has been proposed in the context of refining incomplete RGB-D data \cite{li2015database} or finding similar models in industrial data sets for initial estimates \cite{bey2011reconstruction}.
Using 2D shapes in conjunction with known 3D object models has been proposed in the context of pose-estimation tasks but usually without the goal of achieving a full 3D object reconstruction of the object or scene. In \cite{zhu2014single}, object silhouettes were used for estimating the 6D pose by training trained a deformable parts model based on a Conditional Random Field for different viewpoints and refining the results using dynamic programming. 

\subsection{Pose estimation based on deep learning}
In the last years, many approaches for pose estimation based on deep learning have been proposed \cite{ssd,posecnn,dope,peng2019pvnet,park2019pix2pose} and evaluated using data sets of the BOP challenge \cite{hodan2018bop,hodan2020bop}. One of the highest-ranking approaches is PoseCNN \cite{posecnn}. It has in common with our approach that it computes object pose directly from RGB-images. The architecture of PoseCNN contains deep convolutional layer to first extract image features. Embedded in the architecture are further processing stages: Additional layers compute semantic labels and the 3D translation parameters of the objects from the features. These information are combined to create bounding boxes and used together with the image features to find the 3D rotation parameters using fully-connected layers. To train the network, 6D poses have to be provided with the image data. For the OccludedLINEMOD data, a mean 6D-pose-estimation accuracy of $24.9$ is reported. Using depth information available with the test data, the 6D pose is refined using the ICP algorithm, and an accuracy of $78.0$ is achieved. Another high-ranking method is DOPE \cite{dope}. A deep network computes belief maps directly from RGB images from which 3D vertices of bounding boxes are computed. Using further processing, the 6D pose of the object is estimated. For training, 6D-pose information is required, and the YCB data set was used for this purpose. To perform robotic experiments, YCB objects \cite{ycb} were purchased to match the training data. Success rates between $58,3\%$ and $91,2\%$ were reported (see also Table~\ref{tab:robot}). By using domain randomization, training could be performed on synthetic data, and the problem of 3D data acquisition for fine-tuning the method with real data could be avoided. The most important difference of the described method to our approach is that they require training data that includes 6D-pose information, while our method requires only labeled RGB images for training.

\section{Method}
\label{method}
\begin{figure*}
    \centering
    \includegraphics[width=0.95\linewidth]{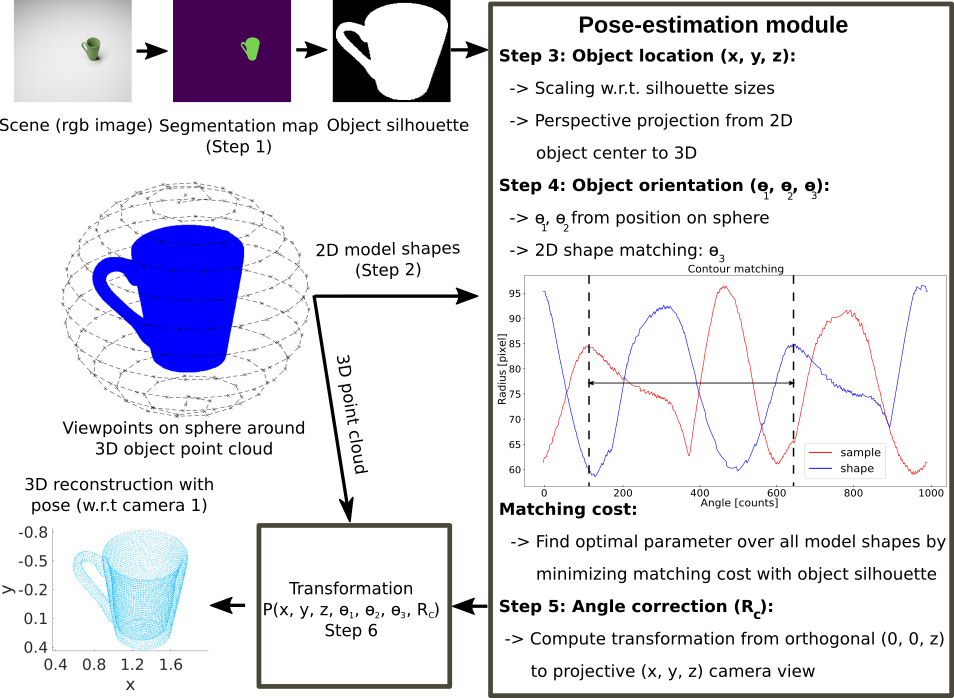}
    \caption{Pipeline of the method. The RGB image is segmented using a deep-learning method. Using 2D shape information and comparing the object silhouette with a data set of 200 previously generated 2D shapes for the object class, pose parameters are estimated in the pose-estimation module. Effects of the perspective projection are corrected with an additional transformation $\mathbf{R_c}$. Using the object 6D pose w.r.t. the camera, a 3D reconstruction is obtained in camera space. 
    }
    \label{fig:pipeline}
\end{figure*}
Our method aims to obtain a full 3D reconstruction of objects from 2D images. During this process, the pose of the object, i.e., the object location ($x$, $y$, $z$) and orientation ($\theta_1$, $\theta_2$, $\theta_3$), is estimated from the RGB images. The computer-vision pipeline consists of the following steps (see also Fig.~\ref{fig:pipeline}):
\newcounter{Lcount}
\begin{list}{(\arabic{Lcount})}
    {\usecounter{Lcount}
    \leftmargin=13pt\listparindent=0pt}
  \item First, the 2D images acquired by the two cameras of the system (one of them is optional) (see Fig.~\ref{fig:schematic}) are segmented and labeled using a deep-learning approach.
  \item Using the class label from the segmentation, the 3D object model (point cloud) of the respective class is drawn from the database together with 2D shapes of the object model obtained for various views sampled from a sphere around the object. Each shape is associated with two angles ($\theta_1$, $\theta_2$) describing the position on the sphere. 
  \item For each possible view, the provisional coordinates of the object $(x,y,z)$ in 3D camera space of camera~1 are estimated from the size of the 2D image segment relative to the size of the model shape using the intrinsic camera parameters and the position of its center in the 2D image.
  \item For each possible view, the 2D silhouette of the model is compared with the 2D silhouette of the 2D segment using a matching approach. This provides also the third angle ($\theta_3$) for the orientation. The match with the lowest cost is selected, providing together with the previous steps the coordinates of the object in 3D camera space and its three orientation angles. 
  \item To account for perspective effects in the 2D projection, the Euler angles are corrected taking the 3D geometry of the problem into account.
  \item Using the parameters estimated in the previous steps, the 3D model point cloud is transformed to the 3D camera space of camera~1, providing a 3D reconstruction of the object. 
  \item To reduce view ambiguities, a second camera is included. For each view processed in steps~3-5, the 2D shape of the object model as seen by camera~2 is generated using the extrinsic matrices of the set-up and the respective pose parameters. The 2D shape is compared with the 2D segment of the second image, replacing the previously computed cost.
\end{list}
Machine learning is only used for the image-segmentation front end of the pipeline (step~1). The remaining steps~ 2-7 do not require any machine learning and thus training data. 

\subsection{Data generation}
\label{method:data}
In this work, we used five different objects: Box, Spoon, Bottle, Cup and Plate. The "Artec Space Spider" laser scanner from the company Artec 3D was used to generate a 3D point clouds for each object, including a texture map, which is useful to render realistic images with software.
The synthetic data set for training the neural network for image segmentation and the scenes for the virtual experiments are generated using the rendering software Blender \cite{blender}. We import the 3D object point clouds for every object class together with its texture map. In the software the objects and cameras can be freely placed and rotated in the scene. The virtual cameras have focal lengths of $16$ millimeters and are placed at location $(-8.95, -9.10, 10.35)$ B.u. (Blender units) with a rotation of $(50^\circ, 0^\circ, -45^\circ)$ for the first camera and the location $(0, 0, 10.10)$ B.u. without rotation for the second camera. We use the included Python-API to write a script that generates random 6D poses in the field of view of both cameras and rotations up to $360^\circ$ for every rotation parameter, applies them to the objects and automatically records images in the resolution of [1024, 736] pixel. We also randomly change the lighting to generalize the data. Additionally, we generate the segmentation maps for the training of the neural network and save the 6D poses for the ground truth. For training, a total number of $10,000$ scenes with $2,000$ scenes per object class were recorded for each camera, resulting in $20,000$ images. For the virtual experiments we record $100$ scenes in total with $20$ scenes per object class. In every scene only one object is placed (see Fig.~ \ref{fig:comparison}). We also acquire a set of RGB images using the camera system from the robotic experiments. This data set is used to further train the neural network. In total, $444$ images of the different objects used in the experiments are taken. 
For the matching procedure, we generate $200$ images for every object model showing different views of the object. The camera views are aligned spherically around the object point cloud (see Fig.~\ref{fig:pipeline}) in a distance of $8$ B.u. for the virtual experiments or $0.8$ meters for the robot experiments. This step is done offline and only once.
\subsection{Object recognition and segmentation}
\label{method:segmentation}
We implement a deep learning method based on ResNet50V2 \cite{resnet} (the first layers are pre-trained on ImageNet \cite{imagenet}) with Tensorflow 2.5 \cite{tensorflow} in Python to do the segmentation. Because ImageNet was originally designed for image classification the top layers are removed and replaced by a series of five 2D-deconvolutional blocks, where every block consist of a 2D-deconvolution layer, batch normalization, and finally ReLU activation. The deconvolutions have a kernel size of 3x3 and 1024, 512, 256, 128 and 64 filters respectively. The backbone of ImageNet decreases the original image resolution from [1024, 736] to [32, 23]. The deconvolutional layers are used to up-sample to the original resolution of [1024, 736]. At the end of the network a Softmax layer is used with a channel size of 6 to label the background and five object classes pixel wise.
We train the newly implemented layers, differing from the ImageNet backbone of the neural network, using mostly synthetic images (see \ref{method:data}). This way we can leverage a bigger data set for the training. Training of the new layers is performed with a learning rate of $10^{-3}$ for $20$ epochs. We also fine-tune the neural network by training all layers, including the backbone, with a very small learning rate of $10^{-5}$ for $30$ epochs. For every training step we use Adam optimization \cite{adam} and a batch size of $2$. To fit the segmentation to the real world experiment setup we use transfer learning. We use the synthetic data set to pre-train the neural network as described above and switch then to a smaller, real-world data set. For the transfer learning we train the neural network with a learning rate of $10^{-4}$ using $20$ epochs. We also fine tune the network with the real-world data, training every layer with a very small learning rate of $10^{-5}$ for 20 epochs. We further use an algorithm for labeling completion that will be described in detail elsewhere to improve the segmentation boundaries of the images obtained during the robot experiments.%
\subsection{3D model selection and model-shape generation}
\label{method:shapes}
The following computations are done in the image coordinate system. After obtaining the segmentation map, we apply a threshold and use a connected components algorithm \cite{bolelli2017two, grana2010optimized} implemented in OpenCV \cite{opencv} to separate every object from the background. Counting the corresponding pixel-wise labels from the original segmentation map determines the object class through a voting approach. The corresponding set of 2D model shapes that have been recorded offline beforehand (see Section~\ref{method:data}) is retrieved. During the matching process, the segment silhouette is only compared with the shapes of this particular set, consisting of $200$ shapes. Identifying the matching shape delivers the parameters used for generating this specific camera view, i.e., two of the three Euler angles $\theta_1$ and $\theta_2$. 

\subsection{3D object location}
\label{method:location}
We determine the coordinates $\mathbf{p}^I=\left( x^I, y^I \right)$ of the center of mass of the segment in the 2D image. Assuming that the model shape is always at the origin of the 2D coordinate system and that the distance $z$ form the camera is known, the respective 3D coordinates can be computed according to $\mathbf{p} = \frac{\mathbf{p}^I \cdot z}{f}$, where $f$ is the camera focal length. The depth $z$ is computed from the relative size of the segment compared and the model shape. Objects further away occupy a smaller region in the 2D image than close objects. For each dimension, a reciprocal relationship $1/z$ applies. For a rectangle, this leads to a factor proportional to $1/z^2$. Since an area of a 2D object can be approximated by a sum of small rectangles, we can assume safely that this factor also applies when dealing with more complicated shapes. Hence, given the size $N_s$ of the segment  and the size $N_m$ of the model shape (both in amount of pixels), we obtain a scaling factor $f_s =\sqrt{\frac{N_s}{N_{m}}} = \frac{z}{z_m}$. Since the distance of the camera to the object model is known, we can now compute the distance of observed object from the camera by rearranging the equation, obtaining $z = z_{m}/f_s$. 
\subsection{2D shape matching}
\label{method:matching}
During the matching process every shape recorded previously is tested to find the best match for the object pose. Every model shape corresponds to an explicit camera view on the sphere, which provides a rotation matrix $\mathbf{R_s}$ fixing two degrees of freedom, $\theta_1$ and $\theta_2$, of the orientation. The remaining parameter $\theta_3$ for the rotation along the camera axis is computed using a 2D silhouette matching between the sample and the model shape.
First, the objects in the sample and shape are cropped and centered to the center of mass. The contours are extracted using the OpenCV implementation of a border tracing algorithm \cite{suzuki1985topological}. Next, the polar coordinates of the object contours are used to obtain parametric curves by describing the distance from the contour point to the center of mass as a function of angle. The parametric curves from the sample and the model shape are brought to equal length using linear interpolation. This allows calculation of the correlation coefficient between both functions and to find the offset for the best fit (see Fig. \ref{fig:pipeline}, step 4). This offset corresponds to the angle $\theta_3$ by which the model shape has to be rotated in 2D to fit the sample. The rotation of the image in 2D corresponds to a 3D rotation around the camera axis with the matrix $\mathbf{R_i}$ using the parameter $\theta_1, \theta_2, \theta_3$.

\subsection{Angle correction}
We compute an additional rotation matrix to correct the perspective effects induced by the perspective projection. An object that is shifted with respect to the camera axis will be seen from a different angle than an object that is located on the camera axis. The rotation matrix $\mathbf{R_c}$ rotates the point cloud around an axis connecting the origin of the camera coordinate system and the center of the object by the angle enclosed by the axis of rotation and the camera axis.

\begin{figure*}
    \centering
    \includegraphics[width=.32\linewidth]{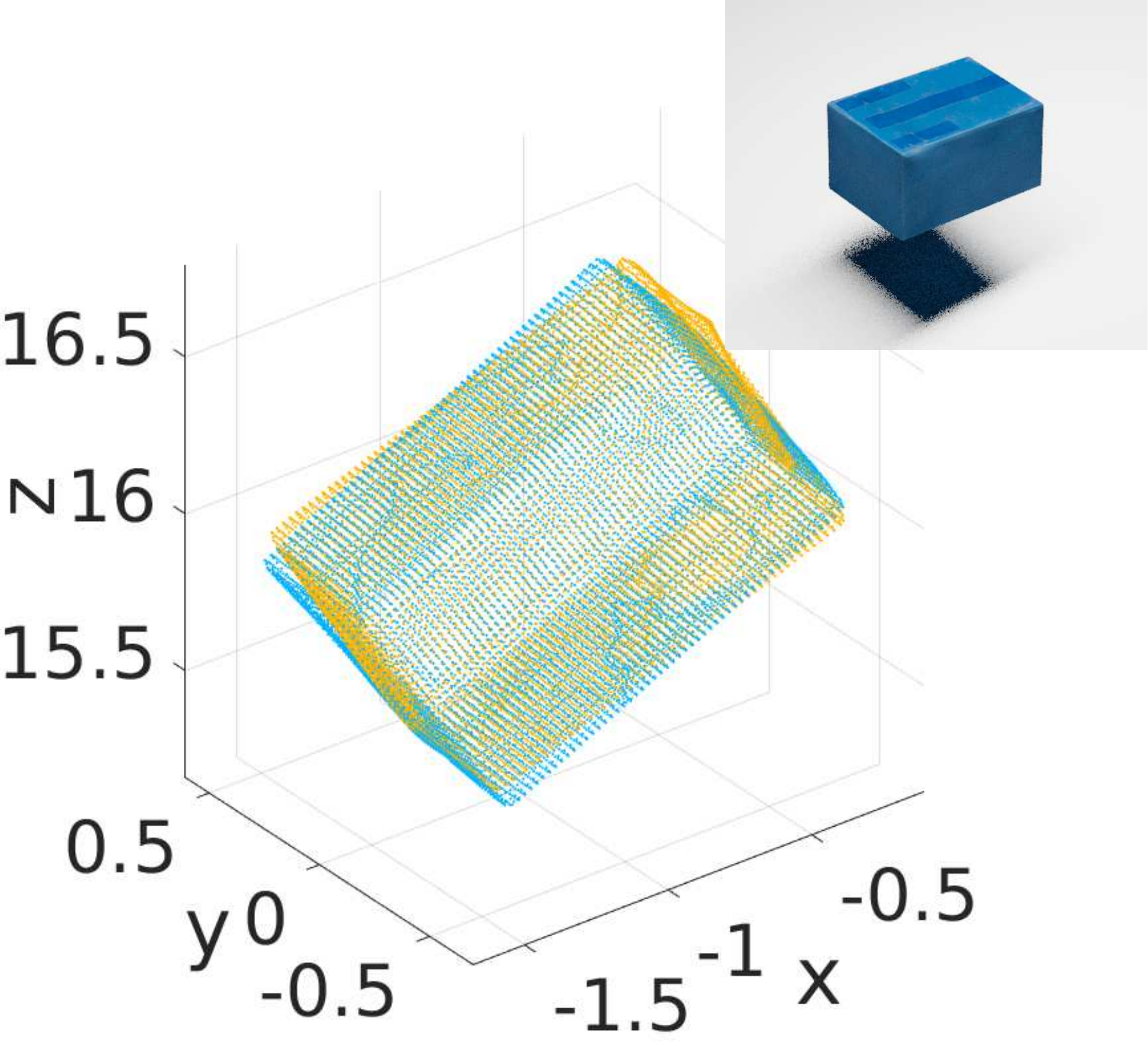}
    \includegraphics[width=.32\linewidth]{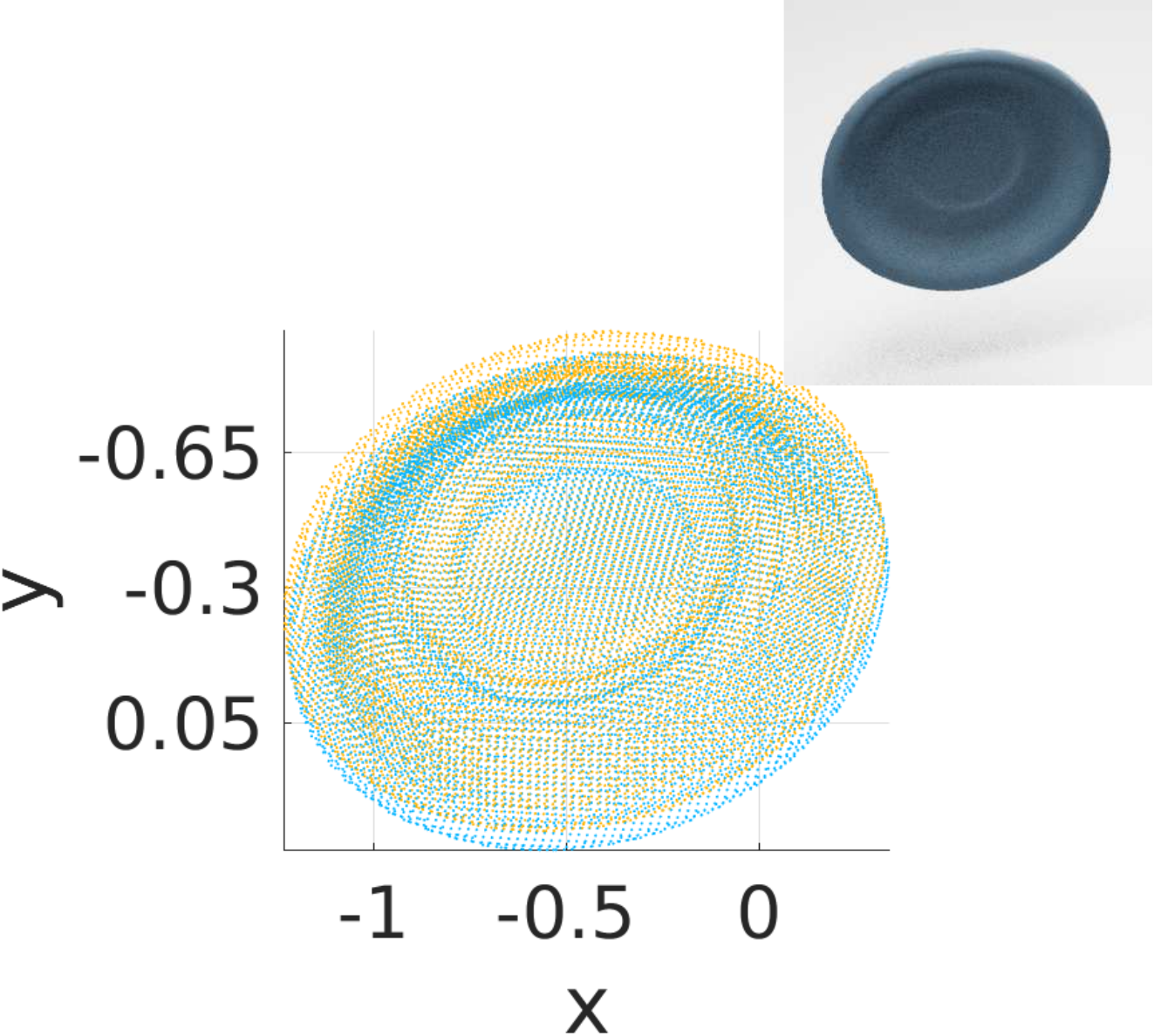}
    \includegraphics[width=.32\linewidth, height=.28\linewidth]{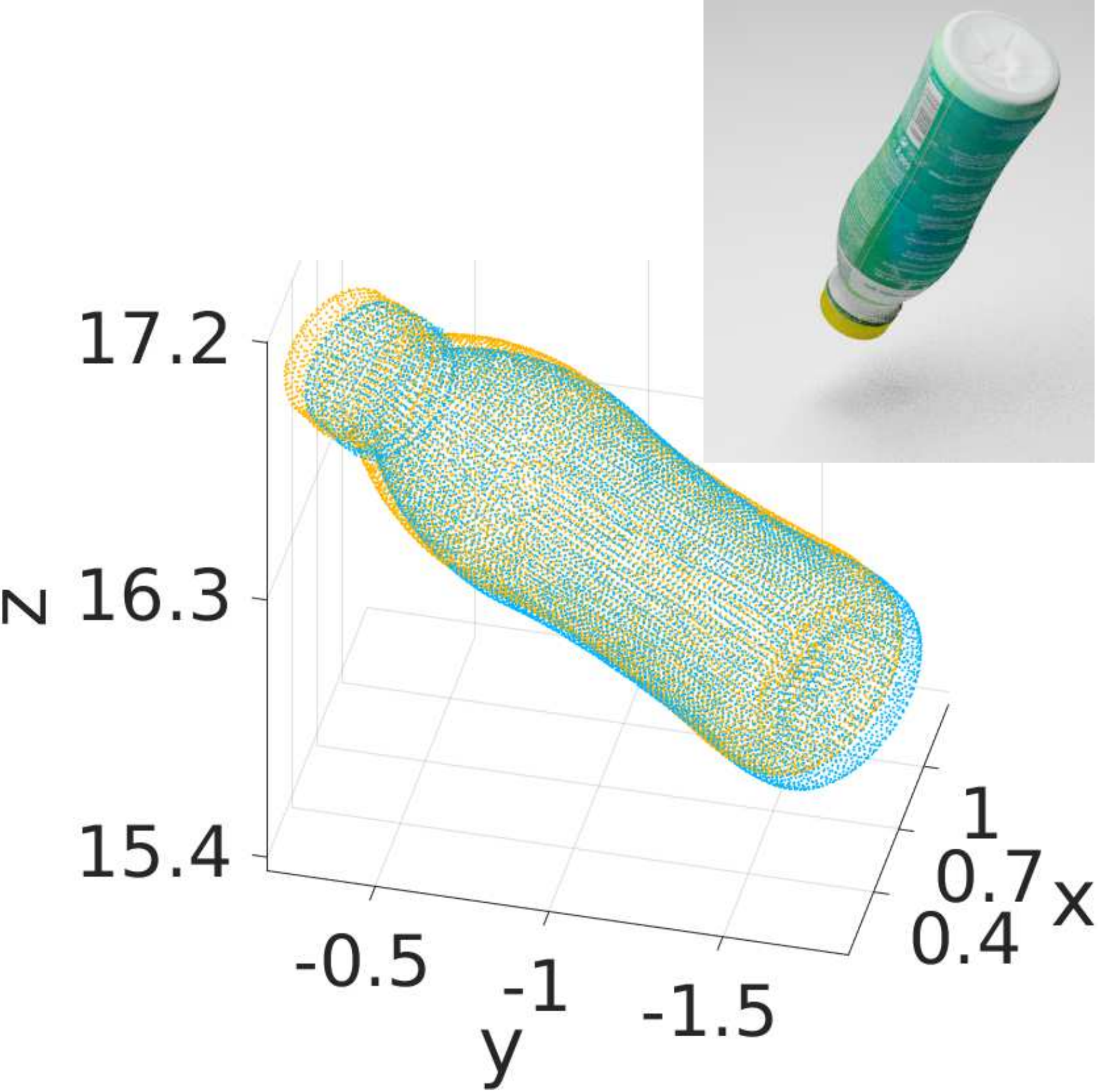}\\
    \includegraphics[width=.28\linewidth]{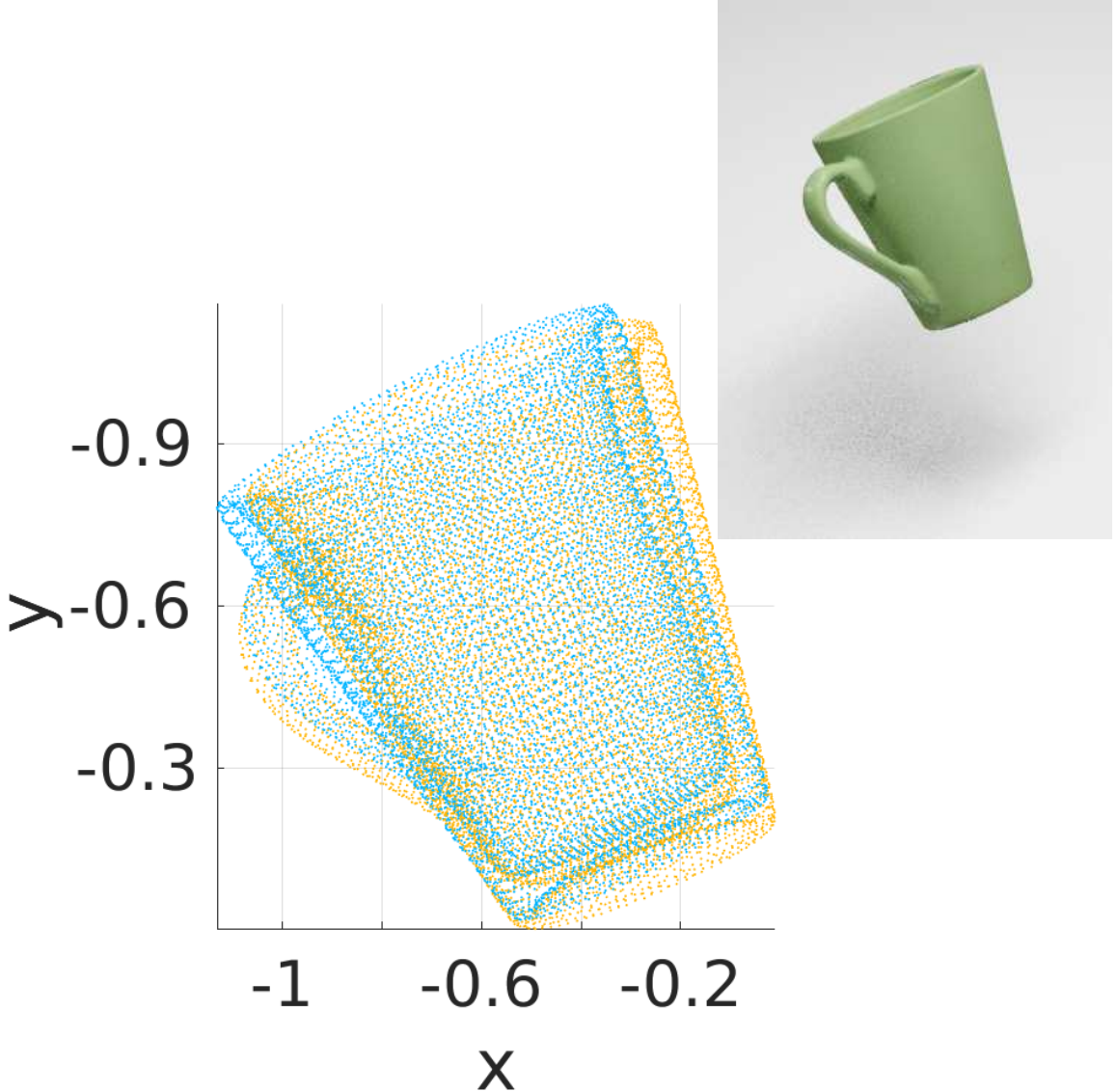}
    \includegraphics[width=.28\linewidth, height=.3\linewidth]{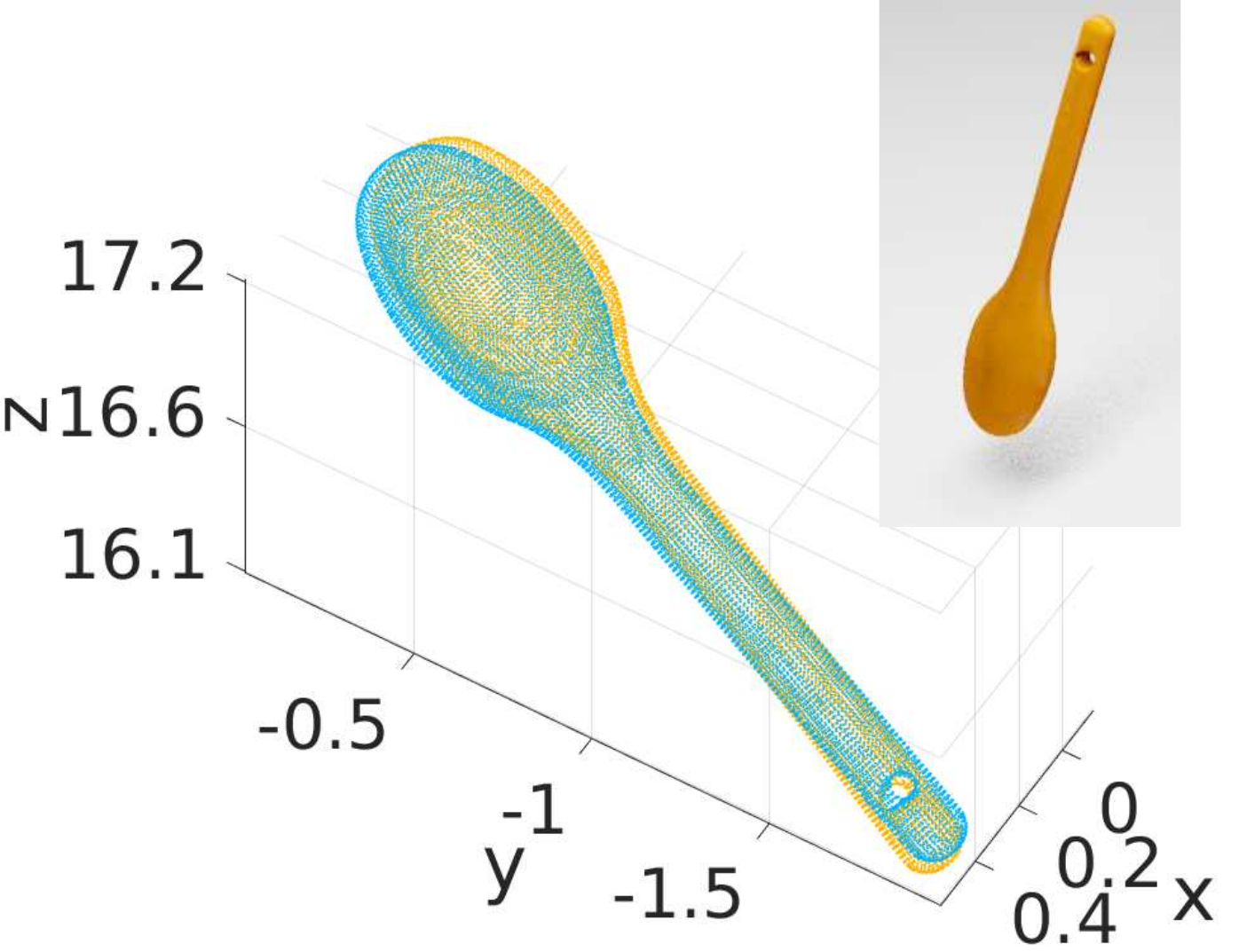}
    \begin{minipage}[b]{.42\linewidth}\scalebox{.9}{
        \begin{tabular}{c|c|c|c|c}
            Object & ADD-S error [$10^{-2}$ B.u.] & \multicolumn{3}{c}{Success rate [\%]}\\
            \hline\hline
             &  & $10$ & $15$ & $20$\\
            \cline{3-5}
            Box & $12.8 \pm 2.2$ & $70$ & $90$ & $95$\\
            Spoon & $68.9 \pm 18.5$ & $20$ & $40$ & $45$\\
            Bottle & $11.4 \pm 1.1$ & $100$ & $100$ & $100$\\
            Cup & $14.3 \pm 1.7$ & $45$ & $75$ & $100$\\
            Plate & $15.6 \pm 2.5$ & $60$ & $80$ & $90$\\
        \end{tabular}}
        \refstepcounter{table}
        \caption*{TABLE \thetable: Results of the simulations. The mean ADD-S error is given in Blender units. Success rates are provided for error thresholds of $10\%$, $15\%$ or $20\%$ with respect to the object point cloud diameter $D$ (see \ref{method:eval_metric}).}
        \label{tab:simulations}
    \end{minipage}
    \caption{Examples of the results of the 3D reconstruction with simulated data using our method (blue point cloud) compared with the ground truth (orange point cloud) for every object class.}
    \label{fig:comparison}
\end{figure*}

\subsection{3D reconstruction}
After obtaining the parameters for the 6D pose, i.e., $x, y, z, \theta_1, \theta_2, \theta_3$, including the matrix $\mathbf{R_c}$ that corrects the perspective effect, we generate transformation matrices in homogeneous coordinates w.r.t. camera coordinate system of the first camera. 
We use the parameter $\theta_1$ and $\theta_2$ to generate the rotation matrix $\mathbf{R_s}$, which corresponds to the viewpoint of the camera on a sphere around the object. $\mathbf{R_i}$ is the rotation matrix for the camera rotation around the optical axis by the angle $\theta_3$. Finally, the translation matrix $\mathbf{T}$ uses the parameters $x, y, z$ for shifting the point cloud. Applying the transformations in the correct order yields a new transformation matrix 
$\mathbf{P} := \mathbf{T} \cdot \mathbf{R_c} \cdot \mathbf{R_i} \cdot \mathbf{R_s}$, describing the 6D pose of the object. We reconstruct the object in the camera space of camera~1 by transforming the model point cloud $M$ according to $\mathbf{q_{cam1}} = \mathbf{P} \cdot \mathbf{q}$, where $\mathbf{q} \in M$ is a point and $\mathbf{q_{cam1}}$ is the transformed point.

\subsection{Second camera}
\label{method:second_cam}
Object symmetries can cause ambiguities in the 6D pose \cite{hinterstoisser2012model, hodavn2016evaluation}. To solve this problem we include a second camera to our method and select the cost of the view of the second camera as the relevant metric for finding the best fit in our method. This cost implicitly contains the information of the first camera. We apply first the inverse extrinsic matrix of the first camera $\mathbf{E_1^{-1}}$ to the reconstructed object point cloud (located in camera space of camera~1) and then the extrinsic matrix $\mathbf{E_2}$ of the second camera, and obtain the coordinates of the point cloud with respect to the second camera, i.e., $\mathbf{q_{cam2}} = \mathbf{E_2} \cdot \mathbf{E_1^{-1}} \cdot \mathbf{q_{cam1}}$.
\subsection{Evaluation metric}
\label{method:eval_metric}
To compare the reconstructed 3D point cloud with the ground truth, we use the error metric ADD-S $:= \frac{1}{m} \sum_{\mathbf{x_1} \in M} \min_{\mathbf{x_2} \in M} \Vert ~ \mathbf{Px_1} - \mathbf{\tilde{P}x_2} ~ \Vert$ proposed by  \cite{hinterstoisser2012model}, where $M$ is the set of all points in the 3D point cloud, $m$ is the number of all points in this set. The computed transformation matrix $\mathbf{P}$ transforms the point cloud to the pose w.r.t the first camera and $\mathbf{\tilde{P}}$ is the ground truth. A pose with an ADD-S error smaller than $15~\%$ of the object point cloud diameter is usually accepted as correct.

\subsection{Robot set up for grasping}
\label{method:path_planning}
The robotic setup consists of a 7-DoF Kuka LWR 4+ arm, a Schunk Dexterous Hand and a computer \cite{robot}. The software for the hand is implemented on ROS and the arm is run via Kuka KRL scripts. The vision system is not integrated for this experiment and the pose estimator outputs are entered manually. The arm is controlled in Cartesian space using the built-in proprietary controller with point-to-point commands. For all objects except plate, only 3D position output from the pose estimator is used, providing the coordinates of the selected grasp point. We have used a fixed orientation parallel to the table.
For the plate, an approach pose, i.e., an offset from the grasp pose in the object's z direction, is added as well. In runs with other objects, only a grasp pose is used. The robot hand has 7-DoF on its three fingers in total. Each finger has a proximal and distal joint, two of them additionally have coupled contrary motion pivoting joints for finger base rotation. The fingers have tactile sensor arrays inside each link. The robot and the tactile sensors are connected to the computer via RS232 connections. In the experiments, we have used two-finger grasps due to convenience. At the start of each run, hand joints are set to the predefined open pose. In grasp command, distal and proximal joints are controlled using a simple feedback control. The joints are run in velocity control mode and the input command is defined by $v^i = (F^i_d - F^i_c) \cdot k$, where $v^i$ is velocity for $i$th joint, $F^i_d$ is desired force, $F^i_c$ is current force read from the sensors, and $k$ is a hand-tuned constant.
The object is assumed grasped when the total pressure on all sensors are above 3 Pa and the joints are commanded to 0 velocity to avoid jittery motion.

\section{Results}
\label{experiments}
\subsection{Quantitative evaluation on synthetic data} To quantitatively evaluate our method, we perform experiments with simulated data (see Section~\ref{method:data} and for examples Fig. \ref{fig:comparison}). The simulations resembled the real-world set up used in the robot experiment. However, for the quantitative evaluation, we want the object poses to be fully diverse without bias towards any degree of freedom. For this reason, objects are not placed ``standing'' on the table. We recorded $100$ scenes with $20$ scenes per object and used our method to estimate the object poses and to reconstruct the objects in 3D camera space and compare them against the ground truth (see Fig.~\ref{fig:comparison}). In the examples shown, the ground truth is a match with the reconstructed point cloud. We further use the ADD-S error to compute the mean absolute distance between the point clouds of our estimation and the ground truth (see Section~\ref{method:eval_metric}). We define thresholds for the ADD-S error compared to the object point cloud diameter $D$ to classify an estimation as correct following the literature (see \cite{hinterstoisser2012model,hodavn2016evaluation}). It basically determines how far the estimated point clouds and ground truth can be apart. We present the percentage of correct estimations for every object class for the thresholds $10~\% \cdot D$, $15~\% \cdot D$ and $20~\% \cdot D$ in Table~\ref{tab:simulations}. The best results are obtained for the bottle object, while the spoon object could only be reconstructed correctly in $40~\%$ of the cases for a standard threshold of $15~\%$. The number of correct cases also dramatically improves for the $15~\%$ and $20~\%$ thresholds relative to $10~\%$. Further experiments (not shown) indicate that the accuracy of the method can be increased by generating more views for every object model or using interpolation. 

\subsection{Quantitative evaluation using Linemod data}
\label{sec:linemod}
The RGB+D dataset Linemod, containing weakly textured objects in cluttered scenes,  is frequently used for evaluating deep-learning based methods for 6D-pose estimation \cite{hodan2018bop,hodan2020bop}. The data sets of the BOP challenge are designed  for benchmarking deep-learning methods that work with bounding boxes, exploit color and texture information, and do not require precise, boundary-preserving image segmentation \cite{posecnn,dope}, but do need 3D depth for training. The data reflects this. Our method on the other hand calculates 6D-pose from precise 2D segment shapes and does not need direct 3D depth. To still enable a comparison, we tested the pose-estimation module (single view), representing the core of our method, by using the ground-truth segments of the data set to calculate 6D-poses directly from segment shape. In Table~\ref{tab:linemod}, columns~4-6, the ADD-S error for the Linemod objects are shown for different thresholds. Our method yields overall better results than PoseCNN \cite{posecnn} if only RGB data is used for testing (see column~2), and not considering differences between Linemod and its extension. 
Importantly though, our method does not require 3D depth for training. A higher accuracy was reported for PoseCNN when 3D depth was used for an ICP-based refinement step (see column~3), but the goal of our work is to compute depth from 2D segments, not to use direct 3D depth as input.

\begin{table}[]
    \centering
    \begin{tabular}{c|c|c|c|c|c}
        Method: & PoseCNN & PoseCNN  & \multicolumn{3}{c}{Our pose-estimation}\\
        & (only RGB) &  + ICP & \multicolumn{3}{c}{module (single view)}\\
        \hline
         Data set:& \scriptsize{OccLinemod} & \scriptsize{OccLinemod} & \multicolumn{3}{c}{\scriptsize{Linemod}}\\
         \hline
         Training data:& \scriptsize{RGB + Depth} & \scriptsize{RGB + Depth} & \multicolumn{3}{c}{\scriptsize{No training}}\\
         \hline
         Input/test data:& \scriptsize{Only RGB} & \scriptsize{RGB + Depth} & \multicolumn{3}{c}{\scriptsize{Image segments}}\\
        \hline
        Threshold [\%]: & $10$ & $10$ & $10$ & $15$ & $20$\\
        \hline \hline
       Object &  \multicolumn{5}{c}{}\\
        \cline{1-6}
        Ape & $9.6$ & $76.2$ & $42.0$ & $62.0$ & $75.0$\\
        Bench Vise & n. a. & n. a. & $65.5$ & $81.0$ & $87.5$\\
        Driller & $41.4$ & $90.3$ & $63.0$ & $81.0$ & $90.0$\\
        Cam & n. a. & n. a. & $74.5$ & $94.0$ & $99.0$\\
        Can & $45.2$ & $87.4$ & $62.0$ & $85.0$ & $91.0$\\
        Iron & n. a. & n. a. & $69.5$ & $87.5$ & $92.0$\\
        Lamp & n. a. & n. a. & $69.5$ & $91.5$ & $97.5$\\
        Phone & n. a. & n. a. & $42.5$ & $69.0$ & $78.0$\\
        Cat & $0.93$ & $52.2$ & $35.0$ & $48.5$ & $58.5$\\
        Holepuncher & $22.1$ & $91.4$ & $48.5$ & $67.0$ & $77.0$\\
        Duck & $19.6$ & $77.7$ & $29.0$ & $67.5$ & $84.5$\\
        Cup & n. a. & n. a. & $74.5$ & $95.0$ & $98.5$\\
        Bowl & n. a. & n. a. & $61.0$ & $71.5$ & $80.5$\\
        Box & $22.0$ & $72.2$ & $83.0$ & $90.5$ & $94.0$\\
        Glue & $38.5$ & $76.7$ & $27.5$ & $44.0$ & $57.5$\\
        \hline
        MEAN & $24.9$ & $78.0$ & $56.4$ & $75.7$ & $84.0$\\
        \hline
    \end{tabular}
    \caption{ADD-S error of our pose-estimation module (single view) for the Linemod data applied to ground-truth segments (see text) without use of depth information for different thresholds (column 4-6).  For comparison, results of PoseCNN \cite{posecnn} for the extended Linemod data set (with occlusions) are shown using only RGB data (2nd column) and using additionally depth information  (3rd column).}
    \label{tab:linemod}
\end{table}

\subsection{Robot experiments} 
\begin{figure}
    \centering
    \includegraphics[width=.99\columnwidth]{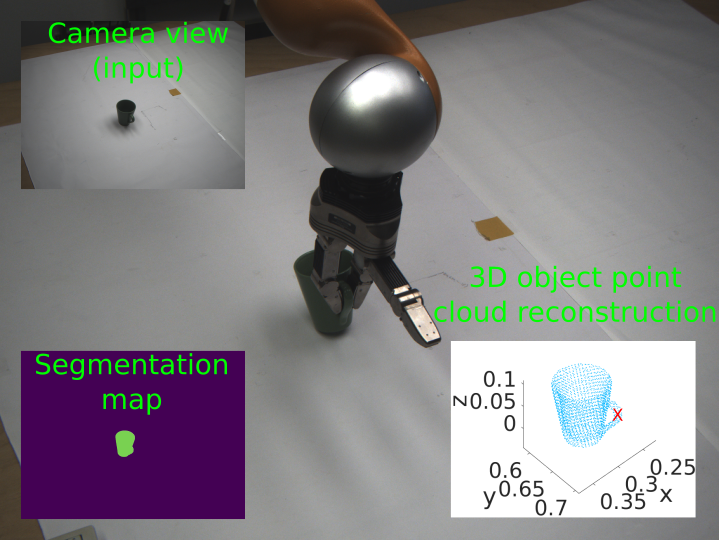}
    \caption{Example of a successful grasp. The reconstruction of the object point cloud in the pose relative to the world coordinate system is used to manually select the grasping point (red cross) for the robot arm.}
    \label{fig:grasp}
\end{figure}

We tested the method in a real-world scenario by using a calibrated camera system of two cameras in a distance of approximately $1$ meter to the table and $1.25$ meter to each other. The intrinsic and extrinsic calibration of the cameras is done by the Anipose calibration routine \cite{Karashchuk2020.05.26.117325} and the calibration of the camera-to-robot coordinate system is done using the Kabsch algorithm on a pre-decided sets of points on the table plane. 
Based on the two RGB images acquired by the two cameras of the set up, the object is being reconstructed in the 3D camera space of camera~1 and a suitable grasp point is manually selected from the point cloud. The coordinates of the grasp point are provided to the path planning and a grasp motion is computed and executed (see Section \ref{method:path_planning}). The goal is to grasp the object at a selected position on the reconstructed point cloud. Fig.~\ref{fig:grasp} shows a successful case for grasping the cup at the specified point on the handle (red marker).

We performed $50$ grasping experiments ($10$ per object) and counted the number of successful grasp to further evaluate our method (see Table~\ref{tab:robot}, top). In accordance with the virtual experiments, the grasps for the spoon object were on average less successful than the ones for the bottle and the plate object. The success rates obtained with our method are in the range of success rates obtained with the deep-learning based method Dope \cite{dope} (see Table~\ref{tab:robot}, bottom). Here, a comparable set of YCB objects in terms of size and shapes was used. The spoon object however can be considered to be more challenging than the objects from the YCB set because of its thin and elongated structures. The respective success rate was thus lower than the others, but it could still be successfully grasped in $4$ out of $10$ cases.

\begin{table}[]
    \centering
    \begin{tabular}{c||c|c|c|c|c}
        \multicolumn{6}{c}{Our approach}\\
        \hline
        Object & Box & Spoon & Bottle & Cup & Plate\\
        \hline
        Success rate [$\%$] & 60 & 40 & 90 & 80 & 100\\
        \hline
    \end{tabular}

    \centering
    \begin{tabular}{c||c|c|c|c|c}
        \multicolumn{6}{c}{}\\
        \multicolumn{6}{c}{DOPE \cite{dope}}\\
        \hline
        YCB & Cracker  & Meat  & Mustard  & Sugar  & Soup\\
        object & box & can & bottle & box & can\\
        \hline
        Success rate [$\%$] & 83.3 & 83.3 & 91.2 & 91.2 & 58.3\\
        \hline
    \end{tabular}
    \caption{Results of grasping experiments with a robot arm. The success rate for ten grasping experiments per object is shown in the upper table. Each experiment corresponded to a new scene presented to the robot system. For comparison, the results of grasping experiments that were reported in \cite{dope} for the method DOPE are shown below. Grasping experiments were repeated $12$ times for each object \cite{dope}.}
    \label{tab:robot}
\end{table}

\section{Conclusion and Discussion}
\label{discussion}
A method for the 3D reconstruction and 6D-pose estimation of known objects given one or two camera images was presented. Using a deep-learning-based approach for image segmentation based on ResNet50V2 \cite{resnet}, objects were classified and segmented in the image. The stored 3D model (point cloud) of the detected object was drawn from a database and 2D shapes were generated for various views of the object. The 2D shapes were then matched with the observed 2D image segment and the remaining parameters of the 6D pose were extracted, allowing reconstruction of the observed object in the 3D camera space. We used a second camera to resolve ambiguities caused by symmetries in the objects. To evaluate the method, a synthetic data set was generated using the rendering software Blender by embedding 3D models of real objects obtained with a 3D scanner. The bottle, box, cup and the plate object scored highest in the virtual experiments with success rates between $90~\%$ and $100~\%$. The corresponding real objects were used in the robotic experiments, and similar scores were obtained except for the box object. The box object is wider than the other objects, which made it difficult to position the robot hand for the grasp. Due to its elongated and thin structure, the spoon was the most challenging object. However, it could still be grasped with reasonable success $4$ out of $10$ times. Our success rates are comparable with the ones of the high-ranking method DOPE \cite{dope} obtained for YCB objects \cite{ycb}. Due to the different requirements on training and test data, data sets from the BOP challenge, which were acquired for testing deep-learning approaches for pose estimation, were of limited use for evaluating our method and obtaining a fair and meaningful comparison. Still, it was possible to apply our pose-estimation module to the Linemod data and compare this component to the method PoseCNN \cite{posecnn}. By design, our method has the advantage that it only requires labeled 2D images for training. Labeled 2D RGB images are comparatively easy to obtain and less costly than 6D-pose information in a real-world scenario. Furthermore, using 2D shapes for pose estimation makes our method independent of features that do not originate from 3D shape, but are correlating with 6D pose for other reasons. 

In the future, we aim at handling multiple objects in a scene, dealing with occlusions by taking scene context into account, and applying the method to dynamic scenes. Since matching scores can be computed independent of each other, an optimized implementation of the code using parallel hardware will contribute to that.

\renewcommand*{\bibfont}{\normalsize\raggedright}
\bibliographystyle{ACM-Reference-Format}
\bibliography{eegbib}

\end{document}